\documentclass[pageno]{jpaper}

\usepackage[normalem]{ulem}
\usepackage{graphicx}
\usepackage{amsmath}
\usepackage{xcolor}
\usepackage{fancyhdr}
\usepackage{hyperref}
\usepackage{bbm}
\usepackage{times}
\usepackage{soul}
\usepackage[utf8]{inputenc}
\usepackage{latexsym}
\usepackage{graphicx}
\usepackage{caption}
\usepackage{amssymb}

\usepackage{algorithm}
\usepackage{algpseudocode}

\usepackage{float}
\floatstyle{plaintop}
\restylefloat{table}

\usepackage{graphicx,epstopdf}
\DeclareGraphicsExtensions{.ps}
\usepackage{stackengine}
\def\delequal{\mathrel{\ensurestackMath{\stackon[1pt]{=}{\scriptstyle\Delta}}}}

\begin{document}

\title{SECS: Efficient Deep Stream Processing via Class Skew Dichotomy}
\author{Boyuan Feng \qquad Kun Wan \qquad Shu Yang \qquad Yufei Ding   \\University of California, Santa Barbara \\ \{boyuan,kun,shuyang1995,yufeiding\}@cs.ucsb.edu }
\date{}
\maketitle

\thispagestyle{empty}

\begin{abstract}
Despite that accelerating convolutional neural network (CNN) receives an increasing research focus, the save on resource consumption always comes with a decrease in accuracy. To both increase accuracy and decrease resource consumption, we explore an environment information, called \textit{class skew}, which is easily available and exists widely in daily life. Since the class skew may switch as time goes, we bring up \textit{probability layer} to utilize class skew without any overhead during the runtime. Further, we observe \textit{class skew dichotomy} that some class skew may appear frequently in the future, called \textit{hot class skew}, and others will never appear again or appear seldom, called \textit{cold class skew}. Inspired by techniques from source code optimization, two modes, i.e., interpretation and compilation, are proposed. The interpretation mode pursues efficient adaption during runtime for cold class skew and the compilation mode aggressively optimize on hot ones for more efficient deployment in the future. Aggressive optimization is processed by class-specific pruning and provides extra benefit. Finally, we design a systematic framework, SECS, to dynamically detect class skew, processing interpretation and compilation, as well as select the most accurate architectures under the runtime resource budget. Extensive evaluations show that SECS can realize end-to-end classification speedups by a factor of $3$x to $11$x relative to state-of-the-art convolutional neural networks, at a higher accuracy.
\end{abstract}

\section{Introduction} \label{Introduction}

Modern convolutional neural networks (CNNs) has made an unprecedented advance in visual recognition tasks. In 2012, AlexNet \cite{krizhevsky2012imagenet} achieved a top-5 error of 17\% on ImageNet \cite{deng2009imagenet}, while previous method could only achieve a top-5 error of 25.7\%. Since then, CNNs have become the dominant method and main research direction in image recognition. In 2015, ResNet \cite{he2016deep} achieved a top-5 error of 3.57\%, suppressing the human-level classification error rate on ImageNet reported as 5.1\% \cite{russakovsky2015imagenet}. The success of CNNs on visual recognition tasks has fueled the desire to deploy these deep networks on various kind of mobile platforms for processing video streams and has an increasingly important role in daily life, e.g., in robotics, self-driving cars, and on cell phones. These mobile platforms usually are memory-constrained and energy-limited while the CNNs are resource-intensive.

To enable the deployment of CNNs on mobile platforms, an increasing research focus has been received by accelerating CNNs, basically trading accuracy for less resource consumption. One approach is to prune the model by reducing the spatial redundancy inside the architecture. LCNN \cite{bagherinezhad2017lcnn} utilizes network quantization to achieve a $5$x speedup at a loss of $7.1$\% accuracy for the ResNet-18 model \cite{he2016deep}. On VGG-16 \cite{simonyan2014very}, filter pruning \cite{lin2018accelerating} is used to reduce computation by $4$x while the accuracy is also decreased by $2.81$\%. Another approach is to build a model store and dynamically select the most accurate model under the available resource budget during runtime. JouleGuard \cite{hoffmann2015jouleguard} utilizes control theory to build a scheduling model and save $3$x computation with a decrease in accuracy of $4$\%. While the resource consumption is reduced, these pruning methods and scheduling models also introduce a decrease in accuracy, which is not desired.

To both reduce resource consumption and increase accuracy, we identify an environment information that has not been studied thoroughly, called \textit{class skew}. Class skew refers to the phenomenon that in an environment, i.e. a specific location or time period, only a few classes may appear while others seldom appear or do not show up at all. For example, only a few people may appear in our lab, even if we may meet thousands of people through the whole year. While a complex model with thousands of classes is required to classify thousands of people, a small model with less than $10$ classes can be sufficient in the lab. Less number of classes indicates a higher accuracy and less resource consumption. For example, if we randomly guess from $1000$ classes, the accuracy is $0.1\%$, while the accuracy would increase to $10\%$ for randomly guessing from $10$ classes. Considering the main constraint of pruning methods is the decrease in accuracy, this increase in accuracy provides more space on optimizing the architecture. Thus more resource consumption can be reduced.

The challenge is how to utilize class skew efficiently, especially considering that class skew may switch frequently as time goes, \textit{e.g.}, every $10$ minutes. It is infeasible to pre-train a sequence of models for each class skew, since there is a mind-bogglingly huge number of class skews. For example, if we take $10$ out of $100$ classes, there would be $1.73 \ast 10^{14}$ combinations of class skew. Existing works \cite{han2016mcdnn, shen2016fast} choose to finetune the model towards the class skew during runtime when class skew switches, based on the technique from transfer learning \cite{doersch2015unsupervised, noroozi2016unsupervised, oquab2014learning, yosinski2014transferable}. With transfer learning, the number of nodes in the last layer will be reduced according to class skew and last few layers will be finetuned by several epochs, which introduce lots of computation overhead and latency. A 14-second or even minutes latency \cite{shen2016fast} will occur every time the class skew switches and the model is adapted. To efficiently adapt the model during runtime towards the class skew, we bring up \textit{probability layer}, an easily-implemented and highly flexible add-on module to existing methods, which introduces no overhead and produces an equivalent or better accuracy than finetuning.

Further examination of class skew reveals the existence of \textit{class skew dichotomy}. Class skew dichotomy represents the phenomenon that some class skews appear frequently in the future, called \textit{hot class skew}, while other class skews never appear again or appear seldom, called \textit{cold class skew}. For example, the hot class skew composed by less-than-ten people from our lab appears everyday when we go to the lab while the cold class skew composed by different types of cats and dogs in a pet house appears generally at most once a month. Inspired by techniques from source code optimization, two modes, called \textit{interpretation} and \textit{compilation}, are designed for cold and hot class skews, respectively. For cold class skew, we utilize the probability layer to efficiently adapt the model during runtime with little or no model optimization, called interpretation mode. Once a class skew appears more frequently than a threshold, we will mark it as hot class skew and optimize the model aggressively by class-specific pruning, called compilation mode. 

\begin{figure}
    \centering
    \includegraphics [scale=0.12] {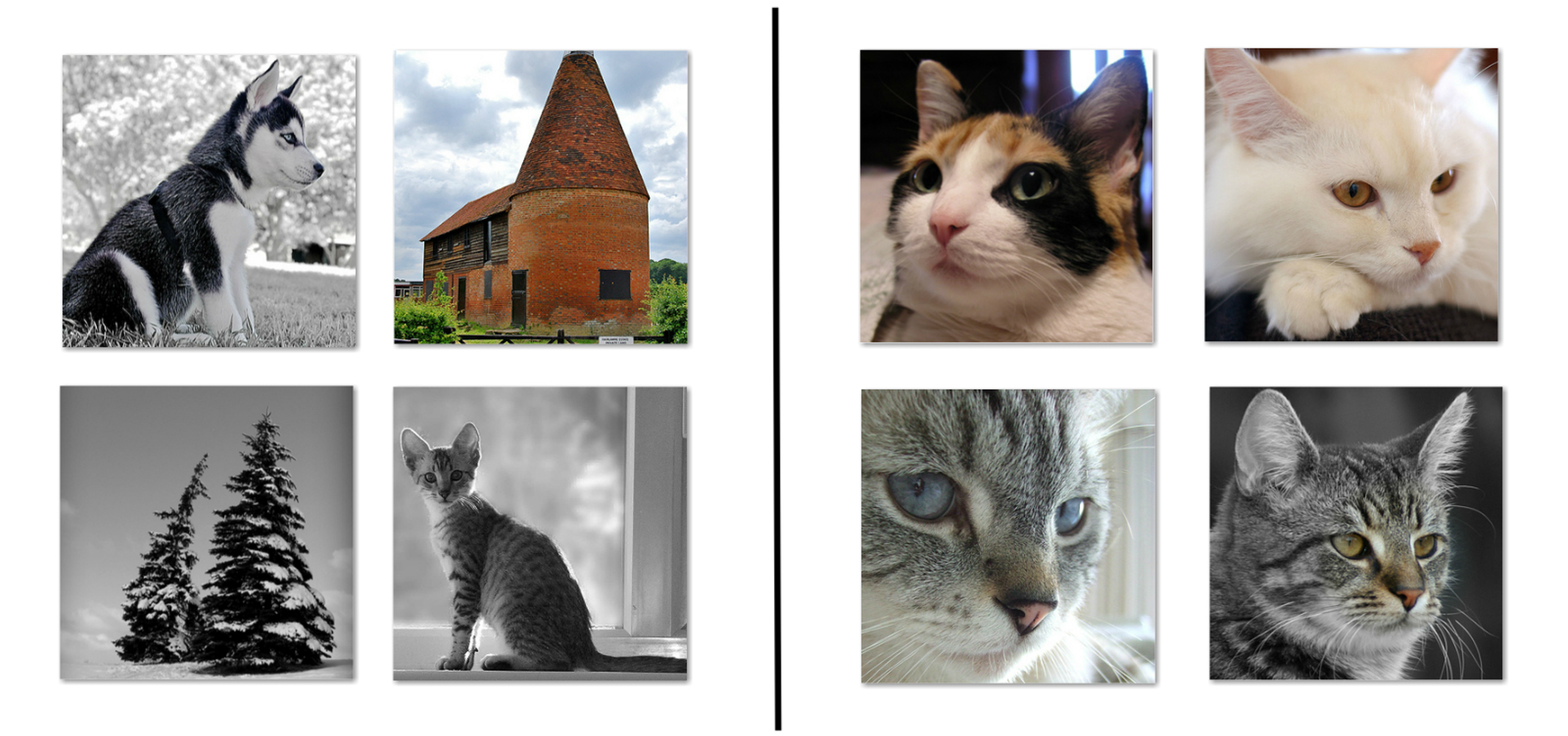}
    \caption{Distinguishing dog, house, tree, cat in the left group is much easier than classifying the four cat types in the right, i.e., kitty cat, tiger cat, Angora cat, and Egyptian cat. }
    \label{fig:classEffect}
\end{figure}

To aggressively optimize the model for hot class skews, class-specific pruning is conducted. While existing works only focus on number of classes, we observe that some classes are easier to distinguish while others are harder, as indicated by Figure \ref{fig:classEffect}.  For example, labeling four more-distinguished classes, \textit{e.g.}, house, cat, dog, and tree, is much easier than classifying four types of cats. Thus, while a complex model is needed to classify four types of cats, a simple model would be enough to label house, cat, dog, and tree. To utilize this intrinsic property of classification target, class-specific pruning is desired, which produces smaller models for simpler class groups by selecting the least number of hyperparameters, i.e., number of layers, channels, and neurons.

Class-specific pruning is challenging because different class skews require different pruned models, especially when we consider the huge number and the frequent switch of class skews. Existing works explore all the combinations of different number of layers, channels, and neurons, to find the Pareto-optimal models consuming least resource for an accuracy. To find the Pareto-optimal models for specific class skews efficiently, we bring up ALPerforation to generate a cascade of models with increasing accuracy, serving as the base for class-specific pruning. With this monotonic cascade of models, we can test the models on class skews in a binary search approach, which only takes logarithm time in number of models in the cascade to generate a class-specific pruned model satisfying the desired accuracy.

In this paper, we present \textbf{SECS}, an efficient deep stream processing service. SECS efficiently detects and utilizes runtime class skew and prunes the candidate model according to the environment and the resource budget, as detailed in section \ref{SECS}. We carefully design a profiler to detect the class skew. Once the class skew is detected, the profiler will further decide whether the class skew is a hot class skew or a cold one. For cold class skews, probability layer is utilized to efficiently adapt the model towards the class skew during runtime with no overhead. If the class skew is marked a hot one, the model will be optimized aggressively with class-specific pruning after the resource constraint is resolved. The adapted models will be stored in the model bank and during runtime a scheduling model will dynamically select the Pareto-optimal models under the available resource budget. Extensive evaluations show that our system can potentially reduce the computation by $3.1$x to $11.2$x times and memory consumption by $3.6$x to $10.6$x times while still producing a higher accuracy.

In summary, we make the following contributions:

\begin{enumerate}
  \item We identify an widely-existing environment information, class skew, which can both decrease resource consumption and increase accuracy.
  \item We bring up \textit{probability layer} to efficiently adapting all models to the class skew without any overhead and achieving an equal or better accuracy than finetuning.
  \item We identify class skew dichotomy and bring up two modes, interpretation and compilation, to adapt the model with no overhead for cold class skews and prune aggressively for hot class skews to gain long-term benefit.
  \item We identify that some class groups are easier to classify while distinguishing some other class groups is harder, and bring up \textit{class-specific pruning} to select hyper-parameters according to whether the targeting classes is easy to classify or not.
  \item We build a system, \textit{SECS}, for efficient deep stream processing service. Extensive evaluations confirm that its benefit is significant.
\end{enumerate}

\section{Motivation} \label{Motivation}
\subsection{Motivating applications}
Among the fastest growing applications, vision-based cognitive assistance applications and robotics visions are two representative categories.

As an example of cognitive assistance applications, smart glasses, Aira \cite{aria2018}, continuously recognize surrounding environment and help the blind person with ordinary tasks, i.e. reading a handwritten note, navigating the grocery store, and even to run the Boston Marathon. To make it feasible, these smart glasses are expected to be lightweight and run at least several hours before recharging. 

On the other hand, robotic visions are expected to recognize objects automatically and work in the wild for days. For example, remote-controlled robotic animals are used by BBC \cite{bbc2018} to search specific animals and document the secret lives of animals in the wild. Teleoperated robots \cite{landmine2018} are used for detecting and removing landmine in various environment. SpotMini \cite{spot-mini} is expected to handle objects, climb stairs, and operate in offices, home, and outdoors.

\paragraph{Common themes.} While these mobile platforms are resource-limited and latency-sensitive, CNN models with high resource consumption are still expected to be deployed on them, since CNNs have a much better performance than traditional approaches. The problem of tension is well- known, and existing methods of solving this problem generally trade accuracy for resource. Now we want to make use of class skew to improve CNN models with both increase in accuracy and decrease in resource consumption. We will discuss the features of class skew and how to utilize it efficiently in details next.

\subsection{Features of input and Opportunities}
The above applications all take input from the lifestyle environment. Such input exhibits class skew, since the activities of mobile devices show strong spatial and temporal locality. 

\paragraph{Temporal locality.} We can view the input stream as a continual camera feed. In the input stream, every frame differs only slightly from previous frames. Thus, the objects in frames usually keep appearing for a time period before the user moves to another scene. For example, a small group of people will appear frequently in a scenario of films, generally lasting for tens of minutes, and another group of people will not appear until the scenario has changed. The class skew appears frequently and provides the chance to simplify CNN models. 

\paragraph{Spatial locality} It is common for human to follow along recurrent trajectories, for example, due to their regular social activities or frequenting a favorite park from time to time. Therefore, there is some level of recurrence of the scenes obtained as part of those activities. Through these repeating scenarios, same class skew will also keep appearing, which make it a feasible choice to optimize CNNs towards the class skew.

The existence of strong temporal and spatial locality indicates the existence of class skew. Experiments on videos of day-to-day life from Youtube \cite{shen2016fast} shows that $10$ objects comprised $90$\% of all objects in $85$\% time. Utilizing class skew will both increase accuracy and decrease resource consumption, as detailed in section \ref{evaluation}. 

We further observe the existence of class skew dichotomy, which means that some class skews appear frequently in the future, called hot class skew, while other class skews never appear again or appear seldom, called cold class skew. Inspired by techniques from source code, this dichotomy motivates us to implement two different levels of optimizations with different overhead on hot and cold class skew, called the compilation and interpretation. Specifically, class-specific pruning can be conducted on hot class skews to achieve extra benefit.

\subsection{Challenges}
In order to utilize class skew more efficiently, we need to deal with class skew switch and explore how to conduct class-specific pruning with low overhead. 

It is obvious that class skew switches frequently as time goes, \textit{e.g.}, every $10$ minutes. Dealing with this phenomenon is hard because it is infeasible to pre-train a model for every class skew during the deployment time, due to the mind-bogglingly huge number of class combinations. In existing works \cite{han2016mcdnn, shen2016fast}, a general model handling all possible classes will be trained at the deployment, and once the class skew is detected during runtime, the model will be adapted correspondingly. Specifically, the number of nodes in the softmax layer will be reduced and the last few layers will be finetuned during runtime towards the class skew under the direction of transfer learning \cite{doersch2015unsupervised, noroozi2016unsupervised, oquab2014learning, yosinski2014transferable}, which consumes energy and memory intensively. 

The challenge of class-specific pruning resides in the obstacle of comparing the performance of various hyper-parameters. Class-specific pruning requires to select the simplest model with least layers, channels, and neurons for a given group of classes. During this process, the relative importance of different layers, channels, and neurons on accuracy needs to be compared and the parameters with the least influence should be removed. However, to get the exact performance, training the new model cannot be avoided while it is unaffordable to do so for each new architecture. Existing papers hand-select architectures, which is not automatical and only test a small space of hyperparameters, i.e. $2$ or $4$ convolutional layers, $32$ or $64$ neurons in each layer, as detailed in section \ref{relatedWork}.

We address these challenges by probability layer, an efficient model adaption algorithm, and ALPerforation, an automatic class-specific pruning method with no finetuning during the process of selection hyper-parameters. Further, we designed a real-time video stream classification framework, SECS, to automatically detect and utilize class skew, as well as select the Pareto-optimal models under the available resource budget.

\section{SECS System Design} \label{SECS}
\subsection{Overview}

\begin{figure} 
\includegraphics[scale=0.095]{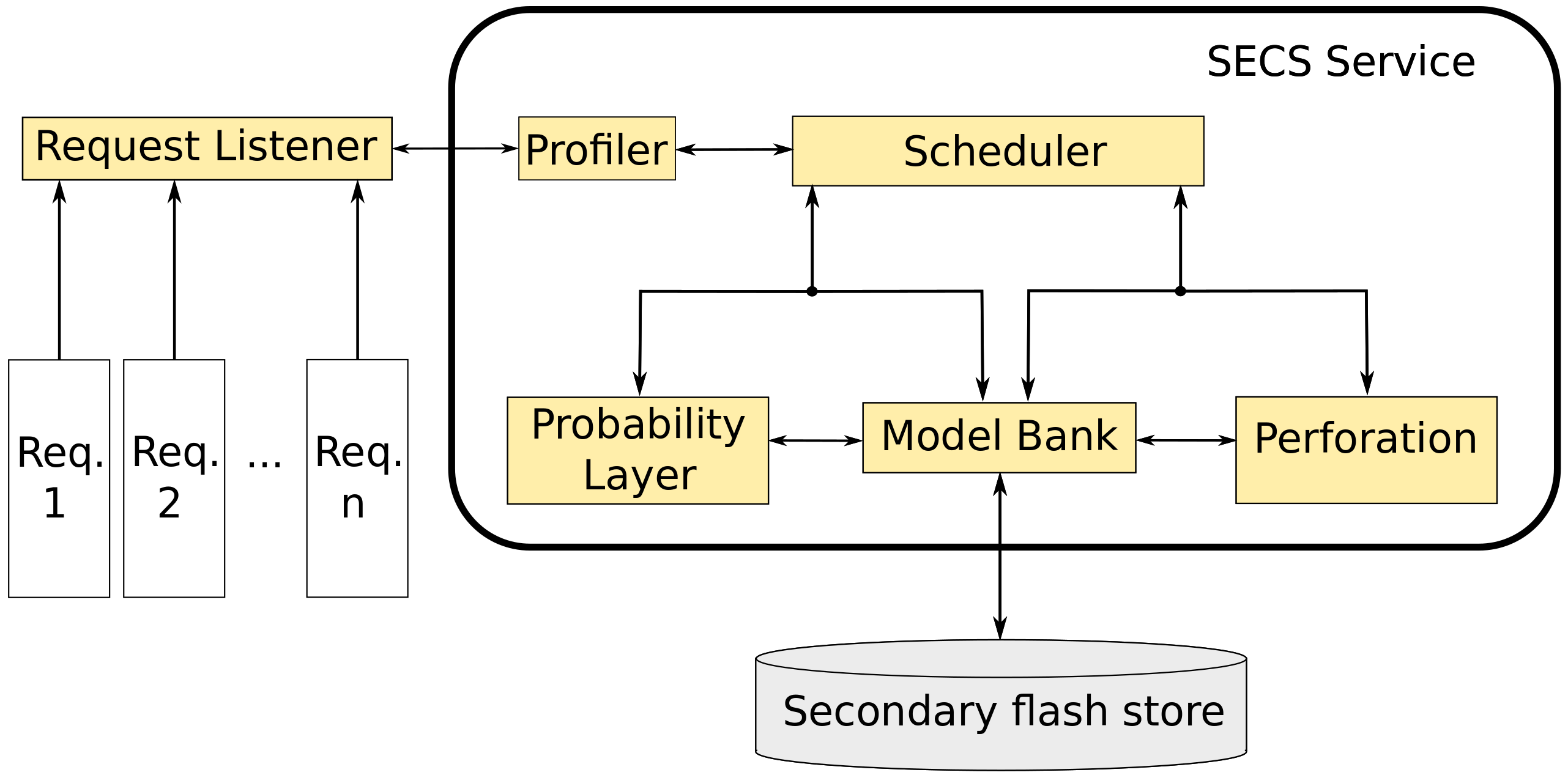}
\caption{System architecture.}
\label{fig:Architecture}
\end{figure}

SECS is a real-time deep stream processing system utilizing class skew efficiently and conducting class specific pruning automatically, as detailed in Figure \ref{fig:Architecture}. The \texttt{RequestListener} maintains a threadpool, summarizes the requests from upper-level requests into a stream, and forwards the stream into \texttt{profiler} for further procedure. During runtime, the \texttt{profiler} maintains current class distribution and effectively detect class skew. The class skew information will be provided to \texttt{scheduler}. Once class skew is detected, the \texttt{scheduler} will either call existing adapted model from \texttt{model bank}, or require the \texttt{probability layer} to efficiently adapt the model. The classification results will be feed back, through \texttt{scheduler}, to \texttt{profiler} which will both respond the \texttt{RequestListener} and update class skew. When \texttt{profiler} detects that a class skew has appeared more frequently than a pre-defined threshold, it will inform the \texttt{scheduler}. Then, during cold time, the \texttt{scheduler} will call \texttt{ALPerforation} to generate class-specific pruned models for more optimized classification in the future. We will discuss the individual steps next.

\paragraph{Notation}
CNN can be viewed as a feed-forward multi-layer architecture that maps the input images $X$ to a vector of estimated probability for each class $\vec{p} = (p_1, p_2, ..., p_n)$, where $n$ is the number of classes and $p_i = P(i|X)$ is the estimated probability $p_i$ for the label $i$ given the input image $X$. In particular, the image feature maps in the $l$-th ($1 \leqslant l \leqslant L$) layer can be denoted by $\mathcal{Z}_l \in \mathbb{R}^{H_l \: \times \: W_l \: \times \: C_l}$, where $H_L$, $W_l$, $C_l$ are the dimensions of the $l$-th feature maps along the axes of sptial height, spatial width, and channels , respectively. $L$ denotes the number of convolutional layers. Individual feature maps in the $l$-th layer could be denoted as $\mathcal{Z}_l^{(k)} \in \mathbb{R}^{H_l \: \times \: W_l}$ with $k \in [1, \:2, \: \dots \:, C_l]$. The individual output feature map $\mathcal{Z}_l^{(k)}$ of the $l$-th convolutional layer is obtained by applying the convolutional operator ($\ast$) to a set of input feature maps with the corresponding filter $\mathcal{W}_l^{(k)} \in \mathbb{R}^{d \: \times \: d \: \times \: C_{l-1}}$, i.e.,
\begin{equation}
    \mathcal{Z}_l^{(k)} = f(\mathcal{Z}_{l-1}^{(k)} \ast \mathcal{W}_l^{(k)}),
\end{equation}
where $f(\cdot)$ is a non-linear activation function. Further, the $l$-th layer can be written as
\begin{equation} \label{eq:1}
    \mathcal{Z}_l = f(\mathcal{Z}_{l-1} \ast \mathcal{W}_l)
\end{equation}
where $\mathcal{W}_l \in \mathbb{R}^{C_l \: \times \:   d \: \times \: d \: \times \: C_{l-1}}$.

\subsection{Probability Layer for interpretation mode}
Interpretation mode is conducted for cold class skews that appear seldomly or never appear again in the future. In the interpretation mode, we target for efficient model adaption since we want to adapt the model during runtime without overhead. Using probability layer, no overhead is introduced in the adaption of general model towards the class skew.

\paragraph{Key Assumption.} 
The main difference between the proposed layer and the original CNNs is that we take the environment information into consideration. In the original CNNs, the prediction for each image will be made individually, assuming a sequence of images is independent and identically distributed (\textit{i.i.d}). However, in real life, this assumption does not hold and strong temporal and spatial locality may exist. Instead, we assume that class skew exists, which means the number of classes and class types are fixed during a time period. Meanwhile, it is still possible that the class skew switches to another class skew after a few minutes, which can be detected and handled by the profiler, as detailed in section \ref{profilerSection}. Further, we assume the existence of a model classifying all possible classes, called the \textit{general model}. This assumption is reasonable because modern CNNs can be trained to classify thousands of classes \cite{krizhevsky2012imagenet, simonyan2014very, szegedy2015going, he2016deep, huang2017densely}.

   \begin{figure}
     \subfloat[Rocket\label{fig:sub2}]{%
       \includegraphics[width=0.23\textwidth]{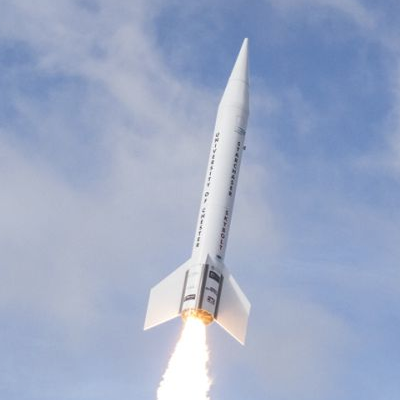}
     }
     \hfill
     \subfloat[Bottle\label{fig:sub3}]{%
       \includegraphics[width=0.23\textwidth]{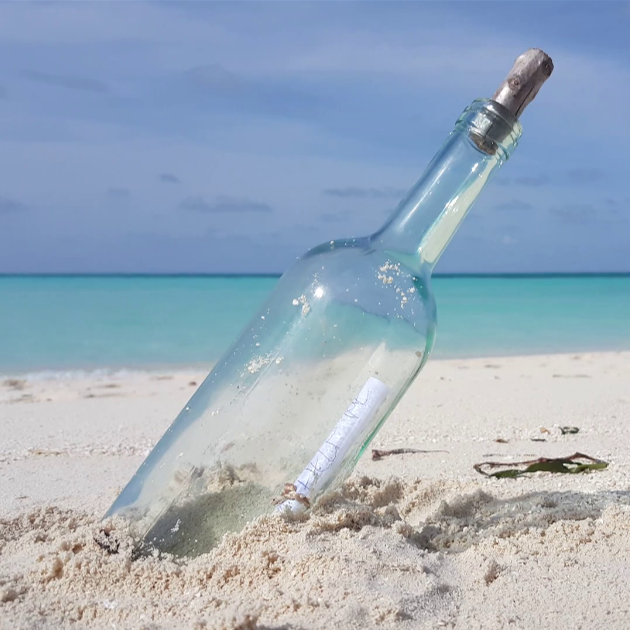}
     }
     \caption{Examples of similar classes.}
     \label{fig:exampleSimilar}
   \end{figure}

\paragraph{Intuition.}
Probability layer helps by using environment information that original CNNs does not use. When human recognizes an object, both vision and environment information will be used, i.e., what we have seen recently and which objects may appear here. However, CNNs can only make use of visual information while discarding environment information, which makes it extremely difficult to distinguish classes with a similar appearance. For example, Figure \ref{fig:sub2} and Figure \ref{fig:sub3} shows images for bottle and rocket respectively. It is hard to distinguish these two classes only from images while environment information can easily rule out rocket in most scenarios. 

\begin{figure}
  \centering
  \includegraphics[width=.45\textwidth]{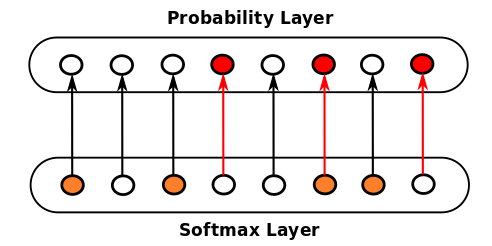}
\caption{Intuition on how probability layer and softmax layer take effect together.}
  \label{fig:intuition}

\end{figure}

Figure \ref{fig:intuition} gives intuition on how probability layer utilize environment information. In Figure \ref{fig:intuition}, the lower row represents the outputs from softmax layer and the upper row represents the probability layer. The orange nodes stand for the classes with high predicted probability in softmax layer and the red nodes stand for the suggestion from the environment. The prediction from probability layer will be selected from the intersection of the set of red nodes and orange nodes, which rules out confusing classes for CNNs. The intersection will not be empty since the red nodes are the classes detected by the general model frequently during the recent time period.

\paragraph{Approach. } Probability layer is an extra layer after the CNN model, rescaling the output of softmax layer. Rescaling is a topic in statistics \cite{saerens2002adjusting}. To the best of our knowledge, we are the first to discuss rescaling in CNN context. The outputs of original CNNs predict the probability for each class and the probability layer will adjust this prediction based on the difference of class distribution in training and testing dataset. In particular, for classes with different distributions in training and testing dataset, the probability layer will rescale the corresponding outputs from softmax layer according to the difference in distribution. For other classes with the same distribution in both training and testing dataset, the outputs of the proposed layer are equal to the outputs of the softmax layer.

The probability layer will take as input the originally predicted probability, class distribution in training dataset, as well as the distribution in testing dataset, and output a vector for the rescaled prediction. The first input is the prediction vector $\mathbf{P}(\cdot|X)$ from the softmax layer, which represents the originally predicted probability for each class from the original CNNs. The second input is a vector of class distribution $\mathbf{P}(\cdot)$ in training dataset and the third one is a vector of class distribution $\mathbf{P}_t(\cdot)$ in testing dataset. The probability layer will rescale the predicted probability in $\mathbf{P}(\cdot|X)$ element wisely and produce as output a vector $\mathbf{P}_t(\cdot|X)$ for the rescaled prediction of each class.

Formally, let the outputs of CNNs with and without rescaling are
\begin{equation}
    P_t(i|X) = \frac{P_t(X|i) \cdot P_t(i)}{P_t(X)}
\end{equation}
and
\begin{equation}
    P(i|X) = \frac{P(X|i) \cdot P(i)}{P(X)}
\end{equation}
respectively. Here $P_t(i)$ means the class distribution in testing dataset and $P_t(i|X)$ represents the predicted probability for class $i$ after the probability layer. We assume that $P_t(X|i)$ equals $P(X|i)$ approximately, where $P(X|i)$ is the distribution of image data for class $i$. This assumption makes sense since, for a class $i$, the selection of input x is random. Through transforming equation $4$ and equation $5$ as well as utilizing $P_t(X|i) = P(X|i)$, we can derive that 
\begin{equation}
    P_t(i|X) = \frac{P_t(i)}{P(i)} \cdot \frac{P(X)}{P_t(X)} \cdot P(i|X).
\end{equation}Considering $\sum_{i=1}^n P_t(i|X) = 1$, we can get the rescaling formular as
\begin{equation}
    P_t(i|X) = \frac{\frac{P_t(i)}{P(i)} \cdot P(i|X)}{\sum_{j=1}^n \frac{P_t(i)}{P(j)} \cdot P(j|X)}
\end{equation}

To give probability layer the ability to detect new classes, we choose not to rescale the outputs from softmax layer when the original model has strong confidence in its prediction and set the formula of probability layer as
\begin{equation} \label{eq: recover}
    P_t(i|X) = \frac{\frac{P_t(i)}{P(i)} \cdot P(i|X)}{\sum_{j=1}^n \frac{P_t(i)}{P(j)} \cdot P(j|X)} \cdot I_{\{P(i|X) < \omega\}} + P(i|X) \cdot I_{\{P(i|X) >= \omega\}}, 
\end{equation}
where $\omega$ is the threshold above which we should trust the original prediction and $I_X$ is the indicator function such that
$I_X(x) \delequal \; $if $x\in X$, return $1$, otherwise return $0$.
If a model has a strong confidence in its prediction, the accuracy would be much higher than the model's average accuracy. Our experiments show that CNNs will give most of the images high predicted probability and the accuracy of these images will exceed average accuracy a lot. Probability layer helps when the original model is confused on the prediction and will not interfere with the decision when the original model has confidence in its prediction.


\subsection{Class-specific pruning for compilation mode}
\textbf{Compilation mode} is conducted for hot class skew that appears frequently in the future. In the compilation mode, we target for extra benefit by selecting the most suitable model according to what compose the class skew and whether the class skew is easy to classify or not, instead of only considering number of classes, \textit{i.e.}, \textbf{class-specific pruning}. To efficiently generate the model bank serving class-specific pruning, we propose to conduct general pruning once for all classes instead of repeating the resource-intensive general pruning techniques for different class skews, since \textit{the best model for different class skews under the same resource constraint is similar}. While the general pruning methods take \textit{exponential time complexity} in number of hyperparameters to find the Pareto-optimal cascade with least resource consumption and highest accuracy, we bring up \textbf{A}ll \textbf{L}evel \textbf{Perforation} (\textit{ALPerforation}) to generate Pareto-optimal cascade in \textit{linear time}. Further, binary search approach is introduced to reduce the linear time complexity to \textit{logarithm time complexity}.

\paragraph{General pruning methods.}

As the base of class-specific pruning, general pruning methods are utilized to generate pruned models, from which the model with targeting accuracy will be selected for a specific class skew. \textit{Exponential time complexity} in number of hyper-parameters is needed to generate all variations with different number of layers, channels, and neurons. Specifically, assuming the number of layers is $N$ and the candidate number of channels for each layer is $16, 32,$ and $64$, the total number of possible variants is $3^N$, which is exponential in the number of layers. However, only a small portion of Pareto-optimal models with the least resource consumption and highest accuracy will be used in the future, while exponential number of combinations are tested and trained. To generate the Pareto-optimal cascade directly without training and testing other models, we bring up ALPerforation to generate Pareto-optimal cascade in \textit{linear time complexity}.

Inspired by techniques from source code optimization, we bring up \textbf{ALPerforation}, which can identify the Pareto-optimal cascade directly with linear time complexity in number of hyper-parameters. ALPerforation globally prunes the unsalient positions, \textit{i.e.}, layers, channels, and neurons, by proposing a global discriminative function on the importance of each position, \textit{i.e.}, influence over the prediction accuracy. With a pre-trained full model, \textit{no training} is required during the process of comparing the importance of different positions, which avoids the exponential time complexity of training and, in \textit{linear time complexity}, generates the desired cascade, denoted as $Cascade_i, \; i \in \{1, 2, \dots, n \}$, where each model is specified by the layer-wise, channel-wise, and neuron-wise perforation rate. Our approach is formulated as following.

\paragraph{Layer-wise perforation.}
For layer-wise perforation, we introduce a global mask $M$ to temporarily mask out unsalient filters in each iteration based on a pretrained model. Therefore, equation \ref{eq:1} can be rewritten as:
\begin{equation} \label{eq: 2}
    \mathcal{Z}_l = M_l \cdot f(\mathcal{Z}_{l_1} \ast \mathcal{W}_l), \ \ \  s.t. \: l = 1,\: 2, \: \dots, \: L,
\end{equation}
where $M_l \in \{0, 1\}$ is a mask with a binary value. $M_l = 1$ if the $l$-th layer is salient, and $0$ otherwise. $\cdot$ denotes the point-wise product. By masking out certain layer as zero, we can view it as skipping the computation of this layer. This operation will make the input to the next layer to be zero and requires finetuning to continue. Instead, we perforate the skipped layer by concatenating same number of channels from previous layers, which can maintain the input tensor size to the next layer and continue prediction without finetuning. Thus equation \ref{eq: 2} can be rewritten as:
\begin{equation}
    \mathcal{Z}_l = g(M_l, f(f(\mathcal{Z}_{l_1} \ast \mathcal{W}_l)), Z_{l-1}), \ \ \  s.t. \: l = 1,\: 2, \: \dots, \: L,
\end{equation}
where $g(x,y,z)  \delequal \; $if $x = 1$, return $y$, otherwise return $z$.

\paragraph{Channel-wise perforation.} Channel-wise perforation could be done in a similar approach with layer-wise perforation. Instead of masking out the whole layer, we adopt a finer granularity with a vector $m_l = \{0,1\}^{C_l}$ as channel-wise mask. The equation \ref{eq:1} can be rewritten as:
\begin{equation} \label{eq: 3}
    \mathcal{Z}_l = f(\mathcal{Z}_{l_1} \ast (m_l \odot \mathcal{W}_l) ), \ \ \  s.t. \: l = 1,\: 2, \: \dots, \: L,
\end{equation}
where $m_l = 1$ if the $l$-th filter is salient, and $0$ otherwise. $\odot$ denotes the channel-wise product. By masking out certain channels as zero, we can view it as skipping the computation of this layer. For the reason of continue computation, we perforate the skipped channels by previous channels. Similar to the heuristic algorithm in layer-wise perforation, we can prune channels in a one-by-one approach.

\paragraph{Neuron-wise perforation. } For neuron-wise perforation, we choose to increase strides to skip neurons instead of using a mask to identify important neurons in each feature maps. The reason is that, as indicated in \cite{aklaghi2018snapea, buckler2018eva, hegde2018ucnn}, the latter process may introduce irregularity and extra overhead in the computation of neurons, which will cancel out the effort in skipping neurons.

\paragraph{Heuristic solution}
A heuristic solution is provided to find the mask. In each round, we will remove the layer with the smallest negative effect on the final accuracy. Each round can be separated into many steps and, in each step, we will remove a single layer while keeping all the other layers. Since ALPerforation is used to recover the feature map size to the layer after the skipped layer, the computation can continue and the final accuracy is still available. Thus in each step, we can get the accuracy after removing the corresponding layer. After iterating through all layers, we can get which layer has the smallest negative effect on the final accuracy and remove this layer.

Our method share some similarities with PCNN
\cite{figurnov2016perforatedcnns} 
in recovering unsalient positions through perforation. Compared to their approach that only prunes neurons, our approach can prune at all three levels, \textit{i.e.}, layer-wise, channel-wise, and neuron-wise, which provides a much larger searching space.

\paragraph{Class-specific pruning.}

The motivation of class-specific pruning comes from the observation that \textit{the best model for different class skews under the same targeting accuracy is different}. Classes with low similarity have much higher testing accuracy than classes with high similarity even if the architecture and the number of classes remain unchanged. Specifically, using a CNN model with 4 convolutional layers and 3 fully connected layers, the accuracy is only 31.49\% to classify \textit{class skew I} (baby, boy, girl, man, and woman) from CIFAR100, while the accuracy would increase to 52.6\% for \textit{class skew II} (bottles, bowls, cans, cups, and plates). Thus, with the same targeting accuracy, less resource is needed to classify a group of classes with high difference.

\begin{figure}
 \subfloat[ALPerforation for class skew I\label{fig:withoutp3}]{%
   \includegraphics[width=0.23\textwidth]{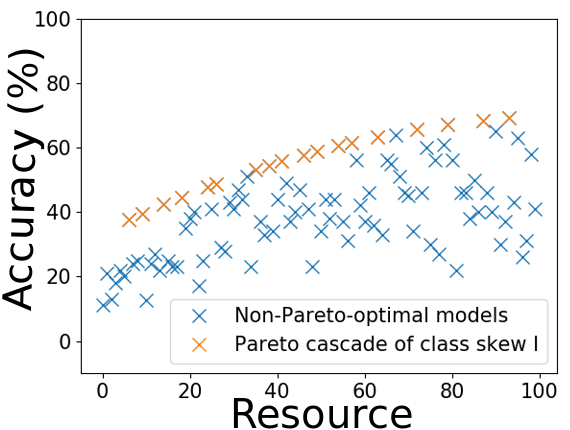}
 }
 \hfill
 \subfloat[ALPerforation for class skew II\label{fig:withp3}]{%
   \includegraphics[width=0.23\textwidth]{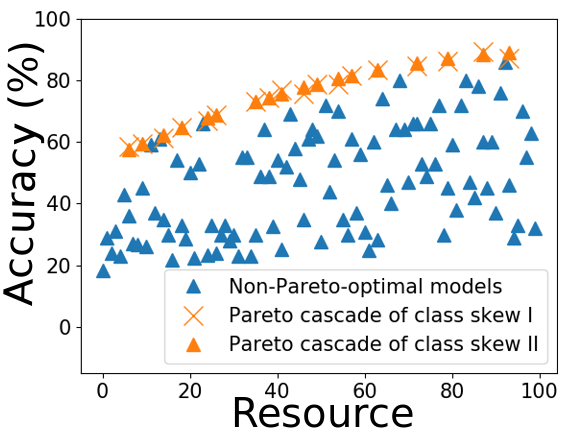}
 }
 \caption{ALPerforation repeated for class skew I and class skew II independently. While each marker denotes a pruned model, the mark X in Figure \ref{fig:withp3} indicates that Pareto-optimal cascade for class skew I has similar performance on class skew II as the cascade for class skew II generated independently. }
 \label{fig:ALPerforation}
\end{figure}

To reducing the overhead of repeating general pruning methods for all class skews, we propose to prune \textbf{once for all class skews}, based on the observation that \textit{best model for different class skews under the same resource constraint is similar}. Specifically, if a model performs better than other models with same resource consumption on a classe skew, it will also achieve the top accuracy among models with similar resource consumption on other class skews, as indicated by Figure 
\ref{fig:ALPerforation}. In Figure \ref{fig:ALPerforation}, 
ALPerforation is conducted twice for class skew I and class skew II, and select the Pareto-optimal cascade of models independently. Evaluting the cascade selected for class skew I on class skew II, we observe that the cascade for class skew I has similar performance with the cascade for class skew II generated by repeating ALPerforation on class skew II independently. This observation motives us to prune once for all class skews in order to conduct class-specific pruning more efficiently.

The insight is that \textit{unsalient positions remain similar for all class skews}. For example, when we repeating the ALPerforation for several class skews on Dense-40  \cite{huang2017densely}
, the positions in latter blocks will be deleted first while the positions in the first block will remain unchanged until all latter blocks have been pruned. This similarity in unsalient positions indicates the similarity in the pruned models for all class skews, which allows us to prune once for all class skews, instead of repeating the ALPerofration for class skews one by one.

\paragraph{Search suitable architectures.}
To find the class-specific pruned model for a class skew, we test the specialzied data following the class skew on the cascade of models and compare the accuracy with our desired accuracy. The naive approach is to test the models one by one, which takes linear time complexity in number of models in the cascade.

To speed up the searching of suitable architectures, we utilize binary search to reduce time complexity from linear time to logrithm time. During the searching process of the simplest model with the targeting accuracy for a specific class skew, we need not to test all models. In fact, the cascade of models generated by ALPerforation will maintain monotonic accuracy. Based on the property of the monotonic accuracy, we can test the models in a binary search approach, which only takes logarithm time.

Specifically, with the sequence of models $Cascade_i, \; i \in \{1, 2, \dots, n \}$ with monotonic accuracy and a targeting accuracy $Acc_{target}$, we can test specialized dataset with the targeting class skew on the model $Cascade_{[n/2]}$ and get the accuracy $acc_{[n/2]}$. If $acc_{[n/2]}$ is below the targeting accuracy $Acc_{target}$, the model $Cascade_{[3n/4]}$ will be tested. Otherwise, the model $Cascade_{[n/4]}$ will be tested. This procedure will continue until we identify a model for the specific class skew with an accuracy close to the targeting accuracy in a threshold $\delta$. 

\subsection{Profiler and scheduler} \label{profilerSection}
Profiler detects class skew and makes a series of decisions on whether or not a class skew exists. The information on class skew will be forwarded to scheduler for further decisions on which model to run and whether the class-specific pruning is necessary or not. We phrase the class skew detection problem as an oracle Bandit problem \cite{auer2002finite, lai1985asymptotically} and utilize WEG algorithm \cite{shen2016fast} to solve it efficiently.

Let denote a stream of images to be classified as $x_1, \: x_2, \: ...,\: x_i, \: ... \in X = \mathbb{R}^n$ and the corresponding true labels as $y_1, \:y_2, \:...,\:y_i, \:... \in Y = [1, ..., k]$. Assume a partition $\pi: I^+ \rightarrow I^+$ over the stream exists, where each partition maintains a distribution $T_{\pi(i})$ and the image $(x_i, y_i)$ is drawn randomly from distribution $T_{\pi(i)}$. Here, the overall series is an abruptly-changing and piece-wise stationary distribution. At test time, neither true labels $y_i$ nor partition $\pi$ is known. Also we do not have any assumptions on how long a stationary distribution exist. The task of profiler is to detect the stationary distribution when it appears and switch to another stationary distribution when it switches.

The emergence and disappearance of class skew can be detected by the WEG algorithm, as shown in the existing work \cite{shen2016fast}. Since the duration of class skew cannot be decided easily, we detect the class skew in a windowed style, as detailed by WEG() in algorithm \ref{alg: algorithm}, line \ref{alg: WEG}. Every $w_{min}$ frames form a window and we can run the full model on each frame and record the distribution in these $w_{min}$ (= 30) frames. We can further compare the record in $S_j$ and $S_{j-1}$. If the difference in appearance times is less than a threshold $\pi_r$ (=2), we conclude that the previous epoch is continuing and use their concatenation as the estimation of class skew.

\begin{algorithm} 
 \small
 \caption{Windowed e-Greedy (WEG)}
  \begin{algorithmic}[1]
    \Function{WEG}{$ $} \label{alg: WEG}
        \For{$i$ in $1, ..., w_{min}$}
            \State $y_t \leftarrow h(t)$
            \State $S_j \leftarrow S_j \oplus [y_t]$
        \EndFor
        \If{$|| S_{j-1}, S_j|| \leq \pi_r$}
            \State $S_j \leftarrow S_{j-1} \oplus S_j$
        \EndIf
        \State \Return $S_j$
    \EndFunction
  
    \Function{Scheduler}{$i, r$}  \Comment{$r$ denotes current class skew} \label{alg: scheduler}
        \State $M \leftarrow collectModel(r)$
        \State $EPF \leftarrow RemainEnergy() / (RemainingTime() \ast F) $
        \State $a, m \leftarrow chooseModel(EPF, r, M)$  
        \State $execute(i,m)$        
    \EndFunction
\end{algorithmic}
 \label{alg: algorithm}

\end{algorithm}

The detected class skew will be processed by the scheduler, as detailed by line \ref{alg: scheduler}. The scheduler will import the class-specific pruned models in the model bank if available. If no class-specific pruned model is available, the scheduler will call the probability layer to adapt the models immediately. These candidate models will serve further decisions along with recorded properties, i.e. energy and memory consumption, as well as accuracy. Then the scheduler estimates the energy per frame by the average of the available resource over remaining times and frames. Finally, $ChooseModel$ will return the most accurate model under the current resource budget.

While original WEG algorithm \cite{shen2016fast} builds a complex scheduling model for the trade-off between accuracy benefit and cost from runtime finetuning, our scheduler does not need it, since the probability layer produces adapted model efficiently and class-specific pruned models are also not produced during runtime. In addition, the detection of the change in class skew can be achieved by equation \ref{eq: recover}, since the probability layer will not interfere with the decision when the original model has confidence in its prediction.

Further, the number of appearance for this class skew will be increased by one and compared with a threshold $\pi_h$. If the class skew has appeared frequently, it will be marked by profiler as the hot class skew and class-specific pruning will be called after the resource constraint on the mobile platform is resolved, i.e. being connected to power and starting to recharge.

\section{Implementation} \label{implementation}
We implement SECS as an end-to-end classification service with Tensorflow \cite{tensorflow2015-whitepaper}.

\subsection{Image classification service}
\paragraph{Probability layer.}
We build probability layer by adding a single extra layer after the original model. The probability layer could be implemented as a variable using \textit{tf.get\_variable} with dimension \textit{1xn}. The probability layer will be initialized with the runtime class skew recorded by \texttt{profiler}, which is freely available and no overhead is introduced. The sum of this variable and the original softmax layer will replace the original softmax layer and used as the classification results. 

\paragraph{ALPerforation.}
There are three levels of ALPerforation, i.e. neuron-wise, channel-wise, and layer-wise. Generally, we speed up computation by skipping several carefully selected positions. The reduced feature map sizes or number of channels will obstacle the forward computation in CNNs. ALPerforation is used to make up the holes that have been skipped using values from adjacent positions. 

Specifically, we implement neuron-wise perforation by increasing strides to reduce computation and use interpolation methods to recover the feature map sizes. By default, \textit{tf.image.resize\_images} and nearest neighbor algorithm is used for interpolation. 

To implement channel-wise perforation, we maintain a global 0-1 mask for all channels indicating whether or not keep the specific channels according to their importance towards accuracy. After masking out the channels, we fill the hole with adjacent channels. \textit{tf.split} can split tensors channel-wise effectively. These splited tensors can be concatenated using \textit{tf.concat} to recover the original number of channels and enable the forward computation with original weights. Layer-wise perforation could be implemented in a similar approach.

\paragraph{Model bank.}
All the generated models are stored in the model bank. Each entry contains the status of a single model, including whether it is a general model or class-specific pruned model, required computation and memory, as well as the accuracy. \textit{N/A} indicates no class skew used in the model. The model bank will support the runtime decision from \texttt{profiler} and only models at the Pareto-optimal bound will be selected for service.

\subsection{Sharing}
It has been identified by existing works \cite{guo2018potluck, jiang2018mainstream} that multiple requests could process the same input stream concurrently. Thus the same classification results could be shared among all requests instead of repeating the computation by each request. To do so, we build an interface \textit{RequestListenser} to process all requests as one and feedback the same results to everyone.

\paragraph{RequestListener.}
As the interface, \texttt{RequestListener} receives messages from applications containing the input images and required accuracy. To synchronize the requests from all applications, the input images received within a time period will be averaged into a single one and all related requests will get the same results. As indicated by \cite{guo2018potluck, jiang2018mainstream}, the choice of this threshold represents a tradeoff between resource efficiency and accuracy. By default, we choose the threshold as a half second.

\subsection{APIs and patches to the application code}
To simplify the usage of SECS service, the user only need to use the interface \textit{requireAccuracy()}, providing required accuracy, and \textit{requireTime()} indicating how long the system are expected to run. Both classification and pruning will be done by SECS service automatically. In addition, an interface, \textit{register()}, is designed for applications to register in the SECS system.

\section{Evaluation} \label{evaluation}

\subsection{General setup}
We implemented probability layer on a state-of-the-art model, DenseNet \cite{huang2017densely}, and evaluated the CNN model with probability layer on specialized datasets with various number of classes and class distribution. We reimplemented the DenseNet on Tensorflow \cite{abadi2016tensorflow}.

\paragraph{Data sets.} While our framework can run in real time, evaluating the recognition performance using existing open source video benchmarks \cite{lomonaco2017core50, jayaraman2016slow, ren2015faster} is difficult, since these benchmarks only focus on a single scene. Various scene switches, although common in real life, are not well presented through these video benchmarks. Therefore, we use synthesized input stream over real-world videos for evaluating our framework. The synthesized input stream is a sequence of images drawn from standard datasets, which is crowd-sourced and well-calibrated. In the generation of the synthesized input stream, various class skews, \textit{e.g.}, number of classes and class distribution, as well as the switches between different class skews, can be emulated, which offer a greater diversity of scenarios and present less favorable (i.e., more challenging) scene sequences for SECS than datasets collected from camera feed. Our results are indicative of SECS's worst-case performance, meaning we can expect better results for real-world videos.

We evaluate SECS on a commonly used image classification datasets, CIFAR100 \cite{krizhevsky2009learning}, from which sequences of images with various class skews are generated. Further, we generate synthetic input streams covering switches between various scenes. 

The CIFAR100 dataset consists of $60,000$ 32x32 color images categorized into $100$ classes, $600$ each. There are $50,000$ training images and $10,000$ test images. To generate a specialized dataset with $n$ classes occupying $p$ percentage, we collect images from these $n$ classes and randomly drawn images from the other $100-n$ classes proportional to the $1-p$ percentage.

\subsection{Interpretation mode for cold class skew}
For class skew that appears first time or seldom, \textit{i.e., cold class skew}, we utilize the probability layer to efficiently adapt the model during runtime with little or no model optimization, called interpretation mode. To evaluate the interpretation mode, we consider two scenarios, strong class skew (p=1) where a few classes occupy all the input stream, and weak class skew (p<1), in which a few classes account for most of the input stream while other classes may still appear. Further, we compare our probability layer with the commonly used runtime model adaption method, \textit{i.e.}, transfer learning.

\begin{figure}[!tbp]
  \centering
  \begin{minipage}[b]{0.235\textwidth}
    \centering
    \includegraphics[width=\textwidth]{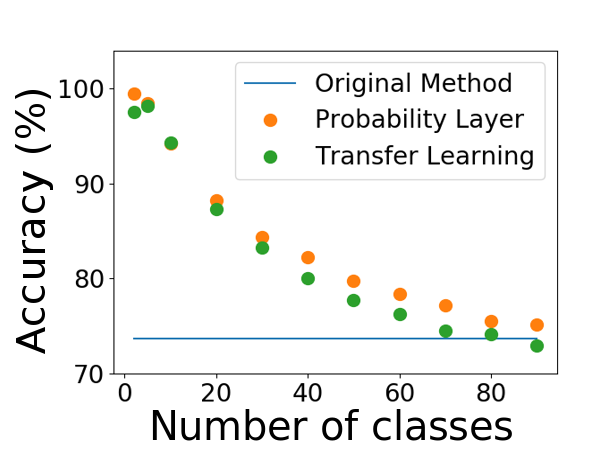}
    \caption{Strong class skew (p=1) with different number of classes}
    \label{fig:PLvsRetrain}
  \end{minipage}
  \hfill
  \begin{minipage}[b]{0.235\textwidth}
    \centering
    \includegraphics[width=\textwidth]{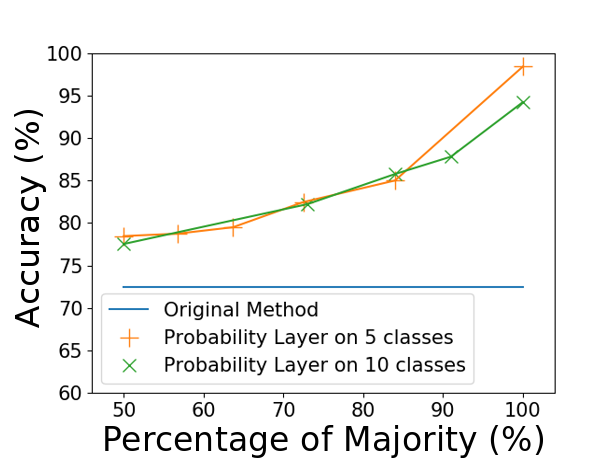}
    \caption{Weak class skew (p<1) with various percentage p}
    \label{fig:variousDistribution}
  \end{minipage}
\end{figure}

\paragraph{Strong class skew (p=1) with different number of classes.}

When the number of classes reduces, benefit can be introduced by both probability layer and transfer learning, as detailed by Figure \ref{fig:PLvsRetrain}. To measure the performance on a class skew with $n$ classes, we randomly sampled $n$ classes for $100$ times and present the average accuracy. We see that significant benefit has been achieved by probability layer for all numbers of classes. When there are 5 classes, more than 20\% increase in accuracy can be achieved without any finetuning. Another point worth noting is that the benefit diminishes slowly as the number of classes increases. Even if there are $40$ classes, a benefit over $10\%$ could still be observed.

Further, Figure \ref{fig:PLvsRetrain} shows that probability layer produces a higher accuracy than transfer learning. Following the published practice \cite{doersch2015unsupervised, han2016mcdnn, oquab2014learning, shen2016fast, yosinski2014transferable}, all fully-connected layers after convolutional layers are finetuned on the generated dataset with same class distribution as the testing dataset for $5$ epochs. For all selected class numbers, probability layer performs better than transfer learning. This advantage of probability layer over transfer learning increases as the number of classes increase. We contribute this phenomenon to the fact that transfer learning may destroy the co-adaption between layers and deteriorate the performance on prediction, as reported in \cite{yosinski2014transferable}. Another point worth noting is that when the number of classes increases over $90$, transfer learning would bring worse accuracy than the original model. In contrast, the probability layer can still bring 2\% advantage over the original model. We believe the reason is that the deterioration of co-adaption between layers leads to a decrease in accuracy and the reduction in the number of classes cannot make up this deterioration when the number of classes is $90$, which is almost same as the original class numbers. Probability layer does not need finetuning and thus avoid this problem. All these observations indicate that probability layer has better ability in using various environment than transfer learning.

Besides accuracy, we should also note that transfer learning during runtime consumes lots of energy.  In each epoch, hundreds of images need to be processed by the mobile platform. As reported in existing works \cite{shen2016fast, han2016mcdnn}, a 14-second or even several-minute latency is required for runtime model adaption using transfer learning. By replacing transfer learning with probability layer, the model can be adapted during runtime without any overhead. 

\paragraph{Weak class skew (p<1) with various percentage p.}
A possible scenario is that $n$ classes only occupy the majority in the input stream while some other classes may also appear. Figure \ref{fig:variousDistribution} shows that probability layer can still have good performance under this scenario. When $10$ classes occupy $90\%$ in the class skew, the accuracy with probability layer can get $87\%$, where a benefit of $15\%$ can be achieved over the original model. As the weight of the $5$ classes decreases, the benefit of probability layer also decreases, since the class skew becomes weaker. However, even if the $5$ classes only occupy $50\%$, a benefit of $5\%$ can still be achieved with probability  layer. Similar results can also be observed for other numbers of classes, \textit{e.g.}, $5$ classes. All these results show that probability layer can bring benefit in various class skew.

\subsection{Compilation mode for hot class skew}
As we recognize that some class skews appear frequently, we will mark it as \textit{hot class skew} and conduct class-specific pruning for these class skew, called the \textit{compilation mode}. To evaluate the compilation mode, we first show that, with class skew, increasing accuracy and decreasing resource consumption can happen simultaneously. Then we proceed to show that two class skews with different class combinations has dramatically different accuracy, which motivates the class-specific pruning.

\begin{table}
    \caption{Both increase accuracy and decrease resource consumption}
    \label{tab:generalPrune}

    \centering
    \begin{tabular}{ c|c|cc } 
     \hline
     Network & Acc. (\%) & Para. (M) & Comp. (M) \\ 
     \hline
     AlexNet \cite{ahmed2016network, NIPS2012_4824} & 56.12 & 57.26 & 302.10 \\
     SqueezeNet \cite{iandola2016squeezenet} & 56.73 & 0.78 & 730.10 \\
     VGG-16 \cite{simonyan2014very} & 71.56 & 133.32 & 797.23 \\
     Dense-40 \cite{huang2017densely} & 72.45 & 1.12 & 264.85 \\
     \hline
     DenseP1 (Ours)& 95.15 & 0.36 (3.1x) & 73.45 (3.6x) \\
     DenseP2 (Ours)& 94.25 & 0.26 (4.3x) & 63.26 (4.2x)\\
     DenseP3 (Ours)& 72.68 & 0.10 (11.2x) & 24.90 (10.6x)\\
     \hline
    \end{tabular}
\end{table}

\paragraph{Both increase accuracy and decrease resource consumption.}
Combining class skew and ALPerforation, we can both increase accuracy and decrease resource consumption, as detailed in Table \ref{tab:generalPrune}. We randomly sampled $5$ classes for $100$ times and present the average accuracy here. On average, the combination of class skews and a sequence of models pruned by ALPerforation leads to more opportunities, \textit{i.e.}, both increase accuracy and decrease resource consumption. Specifically, an increase in accuracy of $22.67\%$ can be achieved with $3.1$x fewer parameters and $3.6$x less computation. By pruning the model aggressively, we can generate model DenseP3, where the accuracy on the class skew is still higher than the accuracy of Dense-40, while consuming $11.2$x fewer parameters and $10.6$x less computation. All of these results indicate that combining class skew with ALPerforation can both increase accuracy and decrease resource consumption. Also, we should note that our approach can achieve a much higher accuracy and dramatically less resource consumption than other architectures that are widely used in practice, \textit{i.e.}, AlexNet \cite{ahmed2016network, NIPS2012_4824}, SqueezeNet \cite{iandola2016squeezenet}, and VGG-16 \cite{simonyan2014very}.

\paragraph{Class-specific pruning}
As indicated by Figure \ref{fig:classEffect} in the introduction, some classes are easier to distinguish while others are hard, which motivates class-specific pruning. As detailed in Table \ref{tab:csPruning}, with the same resource consumption, two class skews with the same number of classes may still have dramatically different accuracy. Specifically, with the same resource consumption, class skew B can have a higher accuracy of $20.28$\% than class skew A. Similarly, class skew D can have a higher accuracy of $13.9$\% than class skew C, even if both class skews have $10$ classes. Through class-specific pruning, class skew B can utilize $11.2$x fewer parameters and $10.6$x less computation to achieve a similar accuracy as class skew A. We can see the same phenomenon in class skew C and D. All these results suggest that class-specific pruning can bring in extra benefit on reducing resource consumption while keeping the performance. Also, we should note that all pruning results have a better accuracy and consume less resource than Dense-40 without utilizing class skews.

\begin{table}
    \caption{Class-specific pruning. Class combinations A, B, C, and D are specified in Table \ref{tab:classGroups}.}
    \label{tab:csPruning}

    \centering
    \begin{tabular}{ c|c|cc } 
     \hline
     Network & Acc. (\%) & Para. (M) & Comp. (M) \\ 
     \hline
     5 classes A & 77.12 & 1.12 (1.0x)& 139.49 (1.96x)\\
     5 classes B & 97.40 & 1.12 (1.0x)& 139.49 (1.96x)\\
     5 classes B & 78.63 & 0.10 (11.2x) & 24.90 (10.6x)\\
     \hline
     10 classes C & 82.90 & 1.12 (1.0x)& 139.49 (1.96x) \\
     10 classes D & 96.80 & 1.12 (1.0x)& 139.49 (1.96x)\\
     10 classes D & 77.80 & 0.10 (11.2x)& 24.90 (10.6x)\\
     \hline
     Dense-40 & 72.45 & 1.12 & 264.85 \\
     \hline
    \end{tabular}
\end{table}

\begin{table}
    \centering
    \begin{tabular}{c|c}
    \hline
     5 classes A & Apple, Baby, Beaver, Boy, Girl \\
     \hline
     5 classes B & Bed, Crab, Bridge, Cloud, Hamster \\
     \hline
     \multirow{2}{*}{10 classes C} & Apple, Baby, Beaver, Boy, Girl, \\
       & Honeybee, Beetle, Maple, Oak, Tank \\
     \hline
     \multirow{2}{*}{10 classes D} & Cloud, Baby, Motorcycle, Tank, Bridge, \\
       & Hamster, Maple, Apple, Crab, Bottle\\
    \hline
    \end{tabular}
    \caption{Randomly selected class groups}
    \label{tab:classGroups}
\end{table}

\subsection{End-to-end evaluation}
We evaluate our end-to-end framework on video streams to measure how well the large speedups of Table \ref{tab:generalPrune} and Table \ref{tab:csPruning} translated to speedups in diverse settings. As an end-to-end framework, our system can be fed with videos to produce classification results by recognizing frames, from which the accuracy is generated as the metric for evaluation. We evaluate our framework on synthetic streams which covers diverse class skews and the switch between consecutive segments.

\paragraph{Overall: Synthetic experiments}
We evaluate our system with synthetically generated stream in order to study diverse settings. For this experiment, we generate a time series of images with various class skews from standard large validation sets of CNNs we use. Each test set comprises of one or two segments where a segment is defined by the number of dominant classes, the skew, and the duration in minutes (=10 minutes, in our implementation). For each segment, we assume that images appear at a fixed interval (1/3 seconds) and that each image is picked from the testing set based on the skew of the segment. For an example of a segment with 5 dominant classes and 90\% skew, we pre-select $5$ classes as dominant classes and pick an image with $90\%$ probability from the dominant classes and an image with $10\%$ probability from the other classes at each time of image arrival over 5 minutes duration. Images in a class are picked in a uniform random way. We also generate traces with two consecutive segments with different configurations to study the effect of moving from one context to the other.

Table \ref{tab:synthe} shows the performance of our system under changing class skews. For segments with no class skew, denoted by "random", our system can achieve the same performance as Dense-40, since no class skew can be detected. When the class skew, $(n=5,p=0.9)$, starts to appear, our system can detect the class skew automatically and select optimized model according to the class skew. An extra accuracy of $2.37\%$ can be achieved while half computation can be saved, due to the benefit of using class skew. When the class skew starts with $(n=20,p=0.9)$ and ends with $(n=10,p=0.8)$, an extra accuracy of $12.4\%$ is achieved consuming $3$x less computations. The evaluation on synthetic streams indicates that our framework can detect class skew automatically and both save resource consumption and increase accuracy when class skew appears.

\begin{table}
    \centering
    \begin{tabular}{c|ccc}
    \hline
    Segments  & Acc. (\%) & Para. (M) & Comp. (M)    \\
    \hline
    (random)      & 72.45 & 1.12 (1.0x) & 264.85 (1.0x) \\
    \hline
    (n=10, p=0.9) & 75.6 & 0.13 (8.61x)   & 26.4 (10.0x)\\
    \hline
    (random)      & \multirow{2}{*}{74.82} & \multirow{2}{*}{0.62 (1.80x)} & \multirow{2}{*}{145.63 (1.81x)}\\
    +(n=5,p=.9) &  & & \\
    \hline
    (n=20,p=.9)      & \multirow{2}{*}{89.65} & \multirow{2}{*}{0.29 (3.86x)}  & \multirow{2}{*}{65.31 (4.07x)} \\
    +(n=10,.8) &  & &\\
    \hline

    \end{tabular}
    \caption{Results on synthesized datasets}
    \label{tab:synthe}
\end{table}

\section{Related Work} \label{relatedWork}
SECS is a real-time deep stream processing system utilizing runtime class skew efficiently and conducting class specific pruning automatically.

\paragraph{Class skew}
Using class skew is an emerging method for both increasing accuracy and decreasing resource consumption \cite{han2016mcdnn, kang2017noscope, shen2016fast}. Our paper is distinguished from existing papers from two main points. First, we bring up probability layer to efficiently adapt the model with no overhead, while existing papers rely on transfer learning to finetune the model towards the class skew during runtime. Specifically, our probability layer introduces no overhead into the runtime model adaption, which fits better for the resource-limited nature on mobile platforms. Secondly, we identify the class skew dichotomy and bring up two modes, interpretation and compilation, to optimize model differently according to whether a class skew is hot or not. Third, we propose class-specific pruning and bring up ALPerforation to efficiently select the smallest model according to whether the targeting class groups is hard to classify or not.

\paragraph{Transfer learning}
Transfer learning has shown benefit in domain adaption and currently is the dominant method, if not the only one, for domain adaption. To solve the problem that the testing dataset is small, we use transfer learning \cite{oquab2014learning}. To use unsupervised dataset \cite{doersch2015unsupervised, noroozi2016unsupervised}, we use transfer learning. Assume we have a large model handling $1000$ classes and in an environment where only $10$ classes appearing, we still use transfer learning \cite{han2016mcdnn, shen2016fast}. Transfer learning has become the only method off the top of the head when we consider the change in classes. Although transfer learning shows various benefit, it also has intrinsic shortage residing in the process of training a CNN model. First, it is hard to decide how many layers we should freeze. Published papers \cite{yosinski2014transferable} reported that freezing fewer layers leads to better performance on different domains and freezing more layers lead to a better performance on similar domains since the co-adapted interactions between layers will be kept. However, it is still hard to decide whether two domains are similar enough and the exact effect of freezing a various number of layers. Second, transfer learning is hard to be conducted unless different settings have been tried. When choosing epoch numbers, it is hard to predict whether the model will converge or collapse after a pre-chosen number of epochs. The choice of learning rate also depends on both model and dataset. Third, the long latency and energy consumption of training a model obstacle the transfer learning on energy-efficient devices, especially in an environment that class number and distributions keep changing. Our method can avoid retraining at all while adapting to the new dataset, thus all the inconvenience related to retraining is avoided naturally.

\paragraph{Model selection.}
To generate a cascade of models with different resource consumption and performance, existing papers utilize hand-selected architectures, which is not an automatic procedure and only a small number of hyperparameters can be tested. Specifically, NoScope \cite{kang2017noscope} only performs model search by varying the number of convolutional layers (2 or 4), number of convolution units in the base layer (32 or 64), and number of neurons in the fully connected layer (32, 64, 128, or 256). MCDNN \cite{han2016mcdnn} chooses between reduce the number of nodes in fully connected layers, decrease the number of kernels, and eliminate convolutional layers entirely. FastVideo \cite{shen2016fast} also manually removes layers, decreases kernel sizes, increases kernel strides, and reduces the size of fully-connected layers, to generate a cascade of models with the tradeoff between accuracy and resource consumption. All these manually selected architectures require human interfere. We observe that the difficulty comes from getting the exact accuracy of each architecture and loose the target to be collecting the relative performance of a sequence of architectures. Further, we bring up ALPerforation to automatically select hyper-parameters according to the targeting classes without any finetuning during the selection process.

\paragraph{Model compression}
Various model compression methods has been brought up, including matrix factorization \cite{jaderberg2014speeding, kim2015compression, romero2014fitnets, xue2014singular}, matrix pruning \cite{chen2015compressing, han2015learning}, and distillation \cite{hinton2015distilling, ba2014deep, dauphin2013big, chen2017learning, lopez2015unifying, kim2015compression,bucilu2006model}. Matrix factorization utilizes the multiplication of two low rank matrices to replace a single high rank matrix. Matrix pruning spasifies matrixes by pruning small digits to be zero. Both directions of model compression do not change the number of nodes in the softmax layer or the architecture. Distillation is another popular pruning method, in which a complex model can teach a smaller model to get a better performance and the smaller model can replace the complex model for the reduction in resource consumption. After we conduct class-specific pruning, all these pruning methods could be utilized to further reduce resource consumption, which will be covered in the future work.

\paragraph{Early stop}
Early stop \cite{teerapittayanon2016branchynet, panda2016conditional} is an architecture with branches and will stop calculating once a branch has enough confidence in that an image has been classified correctly. Early stop contains various architectures to achieve energy efficiency by reducing unnecessary computation. This is orthogonal to runtime specialization and could be included into the model bank.

\paragraph{Reducing resolution}
Reducing resolution has been reported as an effective approach in optimizing architecture \cite{krizhevsky2009learning, fu2017look, howard2017mobilenets}. Reducing resolution indicates the proportional reduction in feature map sizes in all layers and thus both reduces computation and memory consumption proportionally. However, reducing resolution forces the reduction in feature maps sizes in a uniform way across all layers. We argue that reducing uniformly is infeasible since different positions in the architecture have a different degree of redundancy. Instead, ALPerforation could reduce feature map sizes, channel numbers, and layer numbers according to the redundancy in different positions and prune the architecture correspondingly, which would introduce more optimization with less penalty in accuracy. Actually, reducing resolution is equivalent to add a pooling layer before the whole architecture or increase strides in the first layer, which is covered by ALPerforation and could be selected when feasible.

Also note that in MobileNet \cite{howard2017mobilenets}, reducing resolution is utilized by adding an extra hyper-parameter selected by hand, which becomes an obstacle for users who are not familiar with CNN architectures. Instead, ALPerforation selects in an end-to-end automatic way and hide procedures from users.

\section{Conclusion and Future Work} \label{conclusion}
We have presented SECS, a real-time deep stream processing system. We identified the environment class skew, that is easily available and can both increase accuracy and decrease resource consumption. Probability layer is brought up to efficiently utilize the class skew. We further identified the class skew dichotomy. To conduct class-specific pruning for hot class skews, we propose ALPerforation, which can introduce extra benefit based on whether or not a group of classes is easy to classify. 

Looking ahead, we believe there is scope to explore further class skew opportunities. In this paper, we have mainly focused on video classification tasks with CNNs. Actually, our optimizations can be generalized to all the other classification techniques, \textit{e.g.}, MLP and LSTM. In fact, even if it is not a classification task, our optimizations could still be applied as long as parts of the architecture is a CNN model. For example, Fast R-CNN is commonly used for object detection, and part of it is CNN for region classification. Further, we will build open source video benchmarks covering more scenes and various scene switches.


\bibliographystyle{plain}
\bibliography{references}

\end{document}